\documentclass[10pt,twocolumn,letterpaper]{article}

\usepackage{wacv}

\usepackage{graphicx}
\usepackage{tabularx} 
\usepackage{booktabs} 
\usepackage{amsmath} 
\usepackage{siunitx}   
\usepackage{adjustbox} 
\usepackage{tikz}
\usetikzlibrary {arrows.meta,calc,positioning}
\usepackage{wrapfig}
\usepackage{xspace}

\usepackage[dvipsnames]{xcolor}
\usepackage{titletoc} 
\usepackage{longtable}
\usepackage{placeins}
\usepackage{siunitx}
\usepackage{multirow}
\usepackage{tikz}
\usepackage{float}

\usepackage[accsupp]{axessibility} 

\usepackage[resetlabels,labeled]{multibib}
\newcites{Sup}{References - Supplementary}

\usepackage{xcolor}
\newcommand{\be}[1]{
    {\color{Green}#1}
}
\newcommand{\wo}[1]{
    {\color{Red}#1}
}

\definecolor{gold}{RGB}{221, 196, 65}
\definecolor{silver}{RGB}{215, 215, 215}
\definecolor{bronze}{RGB}{126, 66, 5}

\newcommand{\tikzcircle}[2][red,fill=red]{\tikz[baseline=-0.7ex]\draw[#1,radius=#2] (0,0) circle ;}%

\newcommand{\gm}{
    \tikzcircle[gold,fill=gold]{2pt}
}
\newcommand{\sm}{
    \tikzcircle[silver,fill=silver]{2pt}
}
\newcommand{\bm}{
    \tikzcircle[bronze,fill=bronze]{2pt}
}

\newcommand{\methodname}{F-INR\xspace}

\definecolor{wacvblue}{rgb}{0.21,0.49,0.74}
\usepackage[pagebackref,breaklinks,colorlinks,allcolors=wacvblue]{hyperref}

\def\wacvPaperID{1371}
\def\confName{WACV}
\def\confYear{2026}

\title{F-INR: Functional Tensor Decomposition for Implicit Neural Representations}

\author{Sai Karthikeya Vemuri$^*$ $\quad$ Tim Büchner$^*$ $\quad$ Joachim Denzler\\
{\normalsize Computer Vision Group} \\
{\normalsize Friedrich Schiller University Jena, Germany}\\
{\tt\small \{sai.karthikeya.vemuri, tim.buechner, joachim.denzler\}@uni-jena.de}
}

\begin{document}
\maketitle

\begin{NoHyper}
\def\thefootnote{*}\footnotetext{These authors contributed equally to this work.}\def\thefootnote{\arabic{footnote}}
\end{NoHyper}

\begin{abstract}
Implicit Neural Representations (INRs) model signals as continuous, differentiable functions.
However, monolithic INRs scale poorly with data dimensionality, leading to excessive training costs.
We propose F-INR, a framework that addresses this limitation by factorizing a high-dimensional INR into a set of compact, axis-specific sub-networks based on functional tensor decomposition.
These sub-networks learn low-dimensional functional components that are then combined via tensor operations.
This factorization reduces computational complexity while additionally improving representational capacity.
F-INR is both architecture- and decomposition-agnostic.
It integrates with various existing INR backbones (e.g., SIREN, WIRE, FINER, Factor Fields) and tensor formats (e.g., CP, TT, Tucker), offering fine-grained control over the speed-accuracy trade-off via the tensor rank and mode.
Our experiments show F-INR accelerates training by up to $20\times$ and improves fidelity by over \num{6.0} dB PSNR compared to state-of-the-art INRs.
We validate these gains on diverse tasks, including image representation, 3D geometry reconstruction, and neural radiance fields.
We further show F-INR's applicability to scientific computing by modeling complex physics simulations. 
Thus, F-INR provides a scalable, flexible, and efficient framework for high-dimensional signal modeling\footnote{Project page: \url{https://f-inr.github.io}}.
\end{abstract}    
\section{Introduction}
\label{sec:intro}
Implicit Neural Representations (INRs) are continuous, functional representations of discrete signals like images~\cite{stanleykenneth, strumpfler_img, yokota, MIRE_2025_CVPR}, videos~\cite{hao2022implicit, Hao_nerv, zheng2022imavatar, mcgan2023improved}, 3D scenes~\cite{mildenhall2020nerf, hu2023Tri-MipRF, mueller2022instant, Chen2022ECCV, icprainerf}, and geometries~\cite{parkDeepSDFLearningContinuous2019, lin2020sdfsrn}.
Implemented via neural networks, these methods map discrete structured data into a continuous function space, facilitating smooth interpolation.
This general nature promotes progress in architectural design, representation design, and practical applications~\cite{parkDeepSDFLearningContinuous2019, sitzmann2019siren, saragadam2022wire, mildenhall2020nerf, MIRE_2025_CVPR, liu2024finer, zhu2024finer++, chen2023neurbf, chen2023factor, icprainerf}.

\begin{figure}
    \centering
    \includegraphics[width=1.0\linewidth]{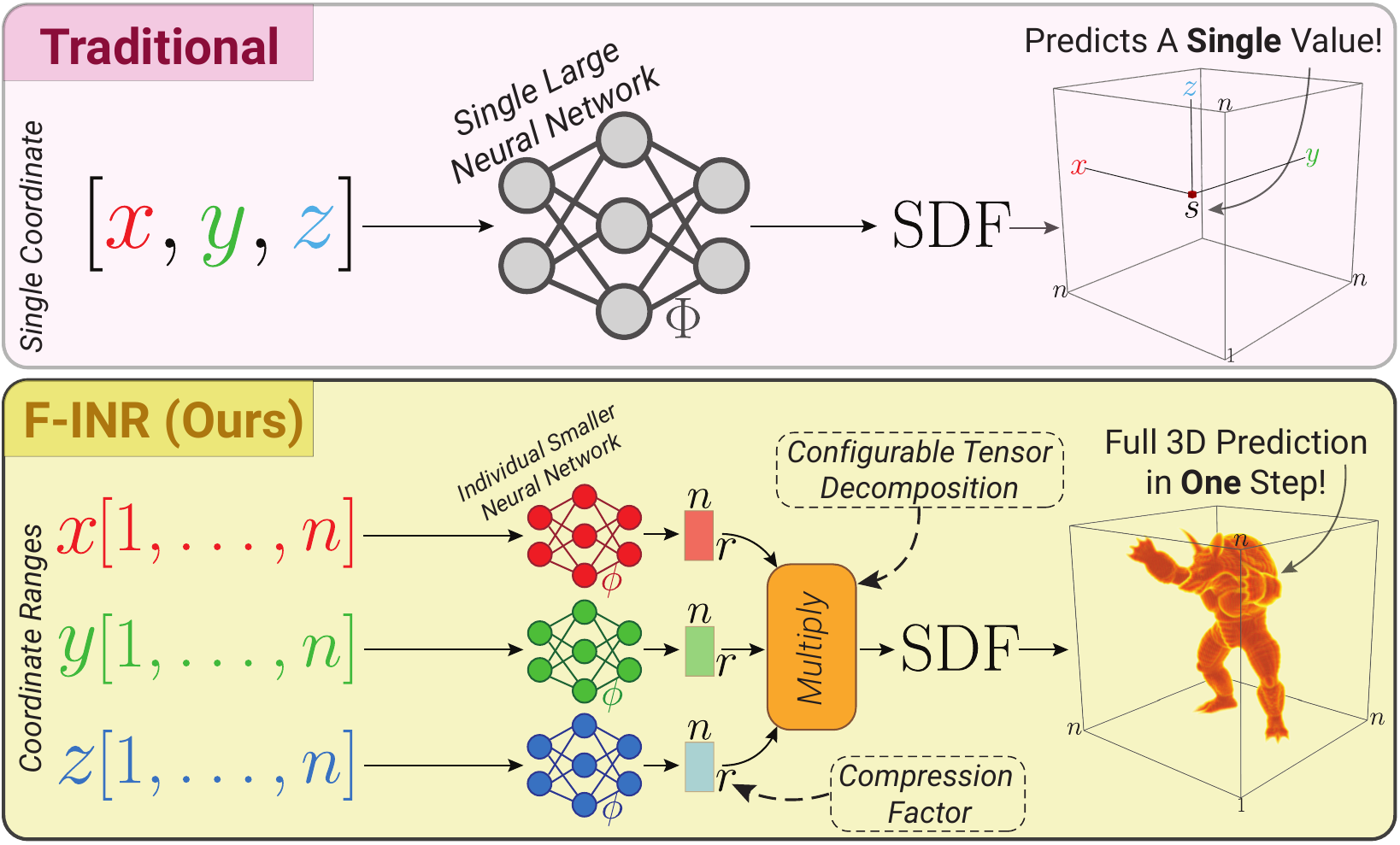}
    \caption{
        \textbf{Efficient INRs via Functional Tensor Decomposition.}
        INR models use a single, large network to predict one value (or a batch of values) at a time.
        Our approach decomposes the function into smaller networks, enabling full prediction in a single step with configurable tensor decomposition modes and compression ranks.
    }
    \vspace{-0.5cm}
    \label{fig:teaser}
\end{figure}

Continuous parametrization offers several advantages over discrete, grid-based representations.
These include higher memory efficiency, the ability to be defined over an unbounded domain, and resolution invariance~\cite{Dupont2022COINNC}.
Specifically, this approach captures fine-grained details, where resolution is determined by the network's capacity and expressiveness rather than the grid~\cite{sitzmann2019siren}.
Additionally, the differentiability of these representations plays a key role in computing gradients and higher-order derivatives using automatic differentiation, relevant for inverse modeling~\cite{sitzmann2019metasdf, sitzmann2019srns, soilsai}.

Another approach for multi-dimensional data is tensor decomposition, used in signal and image analysis~\cite{koldatensordecomps, maolinttn, koldatensor2, zhao, AuBaKo16, BaKlKo20, frosttdataset, SparTen}.
It models high-dimensional signals as combinations of low-rank, low-dimensional components.
Yet, these are confined to discrete grid settings.

Both INR and tensor decomposition have their respective advantages, and combining these could efficiently represent complex data. 
Thus, we propose \textbf{F-INR}, an INR reformulation that leverages the strengths of tensor decompositions.
F-INR uses dedicated univariate neural networks to learn a variably separated form of INR, retaining the benefits of continuous representations by facilitating low-dimensional components.
A general illustration is provided in \Cref{fig:teaser}.
In this study, we focus on combining neural networks with three tensor decomposition techniques, specifically CP~\cite{cphitchcock}, TT~\cite{ttdecomp}, and Tucker~\cite{Tucker1966}.
We establish versatility across various INR tasks, including image encoding, geometry representation via SDFs, and neural radiance fields.
\textbf{F-INR} accelerates existing INRs in training speed (up to $20\times$), improves task-specific metrics, and qualitative results.
Lastly, we test F-INR by modeling physics simulations, showing that existing INR methods can be applied to novel domains.

Our approach demonstrates that factorizing the functional representation unlocks significant improvements, a strategy that directly leverages, enhances, and combines recent architectural advances.
Our work makes the following contributions:
1) We introduce functional tensor decomposition for INRs, a new paradigm for representation learning orthogonal to network design.
2) We propose F-INR, a modular framework implementing this paradigm, agnostic to both the INR backbone and decomposition format.
3) We demonstrate that F-INR enhances existing methods, accelerating training by up to $20\times$ while improving fidelity.

\section{Related Work}
\label{sec:relwork}
\textbf{Implicit Neural Representation (INR)} evolved through several stages. 
Initial ideas introduced explicit coordinate mechanisms, such as positional encoding~\cite{mildenhall2020nerf} and random Fourier feature mappings~\cite{Tancik2020FourierFL}, to enhance the MLP representation capacity.  
Subsequent works focused on activation functions and model architectures:
SIREN introduced fixed sinusoidal activations~\cite{sitzmann2019siren} and FINER made them learnable~\cite{liu2024finer,zhu2024finer++}, WIRE with Gabor wavelet activations~\cite{saragadam2022wire}, and InstantNGP with hash-based encoding~\cite{mueller2022instant}.
Notably, these strongly enhanced the INR expressiveness.
Recent work is application driven, e.g., data compression~\cite{strumpfler_img, hao2022implicit, Dupont2022COINNC}, computer vision~\cite{gropp_igr, parkDeepSDFLearningContinuous2019, Atzmon_2020_CVPR, chibane2020ndf, mcgan2023improved}, graphics~\cite{sitzmann2019srns, DVR, lin2020sdfsrn, yu2020pixelnerf, mildenhall2020nerf, sitzmann2019metasdf}, and robotics~\cite{li20213d, chen2021fullbodyvisualselfmodelingrobot, simeonovdu2021ndf}. 
\newline
\textbf{Tensor decomposition} represents high-dimensional data as compositions of smaller factors.
Classical decompositions split data into mode-wise components~\cite{cphitchcock, koldatensor2, koldatensordecomps, ttdecomp, Tucker1966}.
Early works on continuous tensor decomposition relied on fixed functional bases, such as Fourier or Chebyshev expansions, which limited their expressive power~\cite{HashemiChebfun, yokota, Nikos}.
Modern approaches overcome this by using neural networks as learnable functional bases, where classical components like Tucker factors or SVD vectors are replaced with dedicated networks for significantly higher fidelity~\cite{chientfnn, imaizumi, Chen2024FlexDLDDL, Saragadam2022DeepTensorLT}.
In physics-informed learning, tensor decomposition was used to solve PDEs by splitting the solution into lower-dimensional neural components.
This yielded faster training and higher accuracy~\cite{vemuri2024, vemuri, cho2023separable, JinMIO}.
However, to our knowledge, no prior work unifies INRs with functional tensor decomposition in a general way.
\newline
\textbf{Tensor decompositions for INR} recently emerged as a combination of low-rank tensor factorization with INR-specific applications, such as NeRFs~\cite{mildenhall2020nerf}.
Works like TensoRF factorized the radiance fields into compact low-rank components~\cite{Chen2022ECCV, hu2023Tri-MipRF, yu2021plenoxels}.
This resulted in fast and memory-efficient view synthesis, an effective but domain-specific solution. Moreover, MLPs are not used to learn the components of tensor decomposition directly but only for feature decoding.
Similarly, CoordX employed split MLPs for each coordinate dimension and low-rank representation~\cite{Liang2022CoordXAI}.
They were fused in the deeper layers, omitting different decomposition modes.
Using MLPs to represent low-rank tensor functions (instead of fixed bases) was proposed in works like~\cite{luo2023low, luo2024revisitingnonlocalselfsimilaritycontinuous, ZhaoECCV}.
They represent multidimensional data continuously, achieving state-of-the-art image in-painting and point cloud up-sampling~\cite{luo2023low, luo2024revisitingnonlocalselfsimilaritycontinuous, ZhaoECCV}.
Factor Field~\cite{chen2023factor} proposed a modular paradigm unifying INR learning as a choice of coordinate transformations and field representations.
\methodname moves beyond Factor Fields' fixed rank-1 combination by introducing a principled paradigm for arbitrary-rank tensor decomposition.
\newline
\textbf{\methodname} unifies and generalizes prior methods into a single, modular paradigm.
Akin to CoordX~\cite{Liang2022CoordXAI}, separated MLPs handle dedicated input dimensions, leveraging smaller sub-networks for efficiency.
It incorporates the low-rank structure found in LRTFR~\cite{hu2023Tri-MipRF}, K-Planes~\cite{kplanes_2023}, or TensoRF~\cite{Chen2022ECCV}, reducing redundancy while retaining expressiveness~\cite{vemuri, cho2023separable}.
Unlike these approaches, \methodname is unconstrained to a specific tensor factorization or application domain. 
It supports several decomposition modes and ranks adaptable to the task data structure.
As F-INR is backend-agnostic, it benefits from advances in architectures, such as~\cite{sitzmann2019siren,saragadam2022wire,chen2023factor}, making structured INR representations more modular, scalable, efficient, and versatile for various tasks.
\section{Functional Tensor Decomposition for INRs}
\label{sec:method}
Real-world signals, such as images or videos, must be stored as a discrete grid of values.
These discretized signals, which are $d$-dimensional and $c$-variate, face limitations in resolution and memory.
Neural networks offer a solution by modeling a continuous version of these signals, known as Implicit Neural Representations (INR), which are both resolution-independent and memory-efficient.
Thus, an INR task is a function $\Phi: \mathbb{R}^{d} \mapsto \mathbb{R}^c$, here estimated by a neural network $\Phi_\theta: \mathbb{R}^d \mapsto \mathbb{R}^c$, mapping a coordinate vector $\mathbf{x} = (x_1, \dots, x_d) \in \mathbb{R}^d$ to a $c$-variate output signal.

We instead propose to reformulate the problem approximation as a tensor product of $d$ smaller neural networks: 
\begin{equation}
    \Phi(\mathbf{x}) \approx \bigotimes_{i=1}^{d} \phi_i(x_i; \theta_i),
\end{equation}
where $\phi_i(\cdot)$ denotes a univariate neural network for the $i$-th dimension with learnable parameters $\theta_i$.
Each network produces a component of particular \emph{rank} to restore the original signal via classical tensor decomposition \emph{modes}~\cite{cphitchcock, ttdecomp, Tucker1966}, denoted as $\otimes$.
The decomposition tensors have continuous functions as basis~\cite{vemuri2024}, and the approach is considered functional tensor decomposition.
Hence, our reformulation introduces the idea of functional-INRs (\methodname).

Notably, this approach is model-agnostic, and any INR architecture can be a \emph{backend}, inheriting its advantages~\cite{sitzmann2019siren, saragadam2022wire, mildenhall2020nerf, Tancik2020FourierFL}.
Two more hyperparameters are \emph{mode} and \emph{rank}.
\emph{Mode} refers to the tensor decomposition used. 
In this study, we consider three established \emph{modes} in tensor theory: 
\begin{enumerate}
    \item{
        \textbf{Canonic-Polyadic~(CP)}~\cite{cphitchcock} involves decomposing the $d$-dimensional tensor into $d$-factor matrices of \emph{rank} $R$, visualized in the \Cref{fig:decomp_cp}.
    }
    \item {
        \textbf{Tensor-Train~(TT)}~\cite{ttdecomp} uses $d$-connected chains (trains) of low-\emph{rank} tensors, shown in~\Cref{fig:decomp_tt}.
    }
    \item {
        \textbf{Tucker~(TU)}~\cite{Tucker1966}, like CP, uses factor matrices to decompose a tensor.
        Furthermore, it contains a smaller core tensor $C$ that captures the interactions between the components of each mode, shown in~\Cref{fig:decomp_tu}. 
    }
\end{enumerate}
\noindent
The \emph{rank} determines a model's expressiveness.
This combination of \emph{backends}, \emph{modes}, and \emph{ranks} control fidelity, compression, and performance of a specific task. 

\begin{figure}[t]
    \centering
    \begin{subfigure}[b]{0.28\linewidth}
        \centering
        \resizebox{\textwidth}{!}{%
            \begin{tikzpicture}[
    node distance=0.5cm,
    cross/.style={path picture={
        \draw[black] (path picture bounding box.south west) -- (path picture bounding box.north east);
        \draw[black] (path picture bounding box.north west) -- (path picture bounding box.south east);
    }}
  ]

  \node[circle, draw] (phi1) {$\phi_1$};
  \node[circle, draw, right=0.8cm of phi1] (phi2) {$\phi_2$};
  \node[circle, draw, cross, above=0.75cm of $(phi1)!0.5!(phi2)$] (cross) {};
  \node[circle, draw, above=of cross] (phi3) {$\phi_3$};

  \draw (phi1) -- ++(0,-0.75) node[below] {$x_1$};
  \draw (phi2) -- ++(0,-0.75) node[below] {$x_2$};
  \draw (phi3) -- ++(0, 0.75) node[above] {$x_3$};

  \draw (phi1) -- (cross) node[midway, above left] {\scriptsize $r$};
  \draw (phi2) -- (cross) node[midway, above right] {\scriptsize$r$};
  \draw (phi3) -- (cross) node[midway, left] {\scriptsize$r$};

\end{tikzpicture}
        }%
        \caption{CP~\cite{cphitchcock}}
        \label{fig:decomp_cp}
    \end{subfigure}
    \hfill
    \begin{subfigure}[b]{0.28\linewidth}
        \centering
        \resizebox{\textwidth}{!}{%
            \begin{tikzpicture}[
    node distance=0.5cm,
  ]
  \node[circle, draw] (phi1) {$\phi_1$};
  \node[circle, draw, below=0.5cm of phi1] (phi2) {$\phi_2$};
  \node[circle, draw, below=0.5cm of phi2] (phi3) {$\phi_3$};

  \draw (phi1) -- ++(-0.75,0.0) node[left] {$x_1$};
  \draw (phi2) -- ++(-0.75,0.0) node[left] {$x_2$};
  \draw (phi3) -- ++(-0.75,0.0) node[left] {$x_3$};

  \draw (phi1) -- (phi2) node[midway, right] {\scriptsize $r$};
  \draw (phi2) -- (phi3) node[midway, right] {\scriptsize $r$};

\end{tikzpicture}
        }%
        \caption{TT~\cite{ttdecomp}}
        \label{fig:decomp_tt}
    \end{subfigure}
    \hfill
    \begin{subfigure}[b]{0.28\linewidth}
        \centering
        \resizebox{\textwidth}{!}{%
            \begin{tikzpicture}[
    node distance=0.40cm,
    cross/.style={path picture={
        \draw[black] (path picture bounding box.south west) -- (path picture bounding box.north east);
        \draw[black] (path picture bounding box.north west) -- (path picture bounding box.south east);
    }}
  ]

  \node[circle, draw] (phi1) {$\phi_1$};
  \node[circle, draw, right=0.80cm of phi1] (phi2) {$\phi_2$};
  \node[circle, draw, above=0.70cm of $(phi1)!0.5!(phi2)$] (cross) {$C$};
  \node[circle, draw, above=of cross] (phi3) {$\phi_3$};

  \draw (phi1) -- ++(0,-0.70) node[below] {$x_1$};
  \draw (phi2) -- ++(0,-0.70) node[below] {$x_2$};
  \draw (phi3) -- ++(0, 0.70) node[above] {$x_3$};

  \draw (phi1) -- (cross) node[midway, above left] {\scriptsize $r_1$};
  \draw (phi2) -- (cross) node[midway, above right] {\scriptsize$r_2$};
  \draw (phi3) -- (cross) node[midway, left] {\scriptsize$r_3$};

\end{tikzpicture}
        }%
        \caption{TU~\cite{Tucker1966}}
        \label{fig:decomp_tu}
    \end{subfigure}
    \caption{
        \textbf{Tensor Diagrams for Three Decompositions \subref{fig:decomp_cp}-\subref{fig:decomp_tu}.}
        This schematic~\cite{penrose1971Applications, koldatensordecomps} defines each circle as a component, and spokes determine its dimension. 
        The spokes highlight the decomposition computation and how to recover the original tensor.
        In \methodname, each component is an individual neural network.
        \vspace{-0.5cm}
    }
    \label{fig:TensorDecompositions}
\end{figure}

\subsection{Computational Benefits and Efficiency}
F-INR mitigates the curse of dimensionality by employing an effective separation of variables approach, which is an efficient representation of high-dimensional functions~\cite{GRUNE2021317, HORNIK1989359, yuce2022structured}. 
Recent works have demonstrated that multiple neural networks connected through tensor decomposition forms are universal approximators~\cite{vemuri2024, cho2023separable, JinMIO}.
We provide the formal proof in the supplementary material.
Even if an exact, separable solution does not exist, a sufficiently large rank can approximate the solution, leveraging the universal approximation power of neural networks~\cite{vemuri2024}.
Previous studies~\cite{HORNIK1989359, Poggio2017, GRUNE2021317} have shown that the number of parameters required for function approximation grows exponentially with the dimensionality, hindering neural networks' ability to learn high-dimensional functions.

The reduced trainable parameter count accelerates training speed. 
This efficiency is further improved by a lower forward pass complexity for each decomposition mode, detailed in \Cref{tab:complexity_table}. 
For grid-aligned tasks, the speedup is higher, as a forward pass on an $N^d$ grid requires only $N \cdot d$ evaluations without interpolations. 
F-INR mirrors a divide-and-conquer idea.
Smaller networks learn a low-dimensional representation of univariate functions that aggregate the high-dimensional signal.

F-INR preserves the differentiability of its neural backbones w.r.t.\ the input coordinates.
This provides an advantage over methods based on discrete grids (e.g., TensoRF, iNGP, DVGO), where gradients can be unstable or undefined.
Therefore, we can apply modern backbones to PDE-constrained problems (e.g., physics simulations) that were tractable only with slower, monolithic architectures.

\begin{table}[t]
	\centering
    \def\arraystretch{1.1}%
	\resizebox{\columnwidth}{!}{%
		\begin{tabular}{@{}p{1.5cm} l p{4.5cm}@{}}
			\toprule
			\textbf{Mode}       & \textbf{Complexity}                   & \textbf{Description}                                       \\
			\midrule
			-                            & $\mathcal{O}(m^2 l n^3)$              & A single network with $n^3$ inputs                    \\
            \cmidrule{1-3}
			CP~\cite{cphitchcock}        & $\mathcal{O}(m^2 l n r +  n^2 r^2)$   & Three networks for $n \times r$ factor matrices            \\
			TT~\cite{ttdecomp} & $\mathcal{O}(m^2 l n r^2 +  n^2 r^2)$ & Two $n \times r$ factors and one $r\times n\times r$ decomposition core     \\
			TU~\cite{Tucker1966}     & $\mathcal{O}(m^2 l n r + r n^3)$      & Three $ n \times r $ factors and one $r\times r\times r$ decomposition core \\
			\bottomrule
		\end{tabular}%
	}
	\caption{
        \textbf{Forward Pass Complexity.}
        We assume a grid of $n\times n\times n$, i.e., $n^3$ data points, and tabulate the computations for a neural network ($m$ features, $l$ layers).
		Note that $r \ll m^2l$ and the sequence of the operations is given in the supplementary.
	}
	\label{tab:complexity_table}
\end{table}

\subsection{Architectural and Conceptual Distinctions}
F-INR uses the low-rank, tensor-factor philosophy of methods like TensoRF~\cite{chen2022tensorf} and Factor Fields~\cite{chen2023factor}, but is independent of discrete learnable grids or factor modules.
Unlike TensoRF’s plane and line grids (which require trilinear interpolation) and Plenoxels’ voxel and spherical harmonic lookup tables~\cite{yu2021plenoxels}, we learn fully continuous per-axis neural networks that map each coordinate to an embedding, then fuse via tensor product, removing lookup tables.

Factor Fields~\cite{chen2023factor} unifies many INRs under a Hadamard product, but requires a monolithic architecture and basis/coefficient split; by contrast, our approach is a tensor decomposition with per-mode MLPs, making rank and encoding choices (Hash~\cite{mildenhall2020nerf}, Fourier~\cite{Tancik2020FourierFL}, etc.) explicit and interchangeable. 
Lastly, CoordX uses one monolithic MLP on concatenated coordinates~\cite{Liang2022CoordXAI}; we enforce variable separability throughout multiple small MLPs.
\begin{figure*}[t]
    \includegraphics[width=\textwidth]{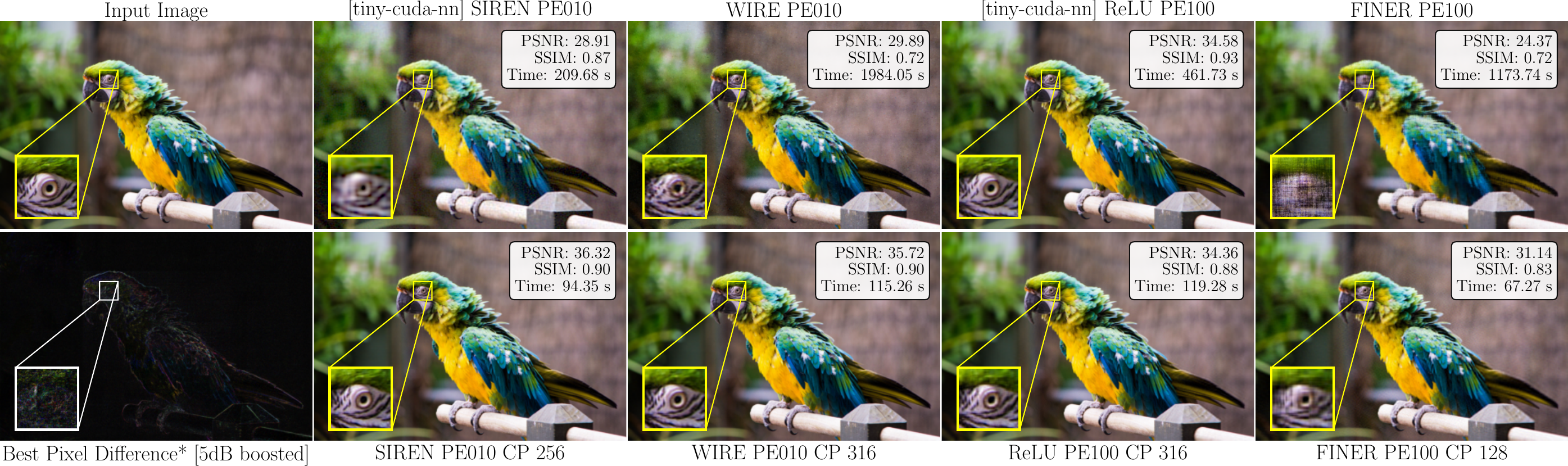}
    \caption{
        \textbf{Qualitative Image Results.}
        F-INR improves both fidelity and training speed over the original backbones. 
        We observe fidelity improvements of up to +7 dB (SIREN) and training speedups reaching $20\times$ for native backbones (WIRE) and $2.2\times$ for optimized CUDA implementations (ReLU).
        Error maps are magnified for visualization.
        See supplementary material for more results.
    }
    \label{fig:img-rep}
\end{figure*}
\section{Experiments}
We explore how combining INR \textbf{backend}, decomposition \textbf{mode}, and tensor \textbf{rank} improves efficiency and fidelity.

Therefore, we focus on foundational backends: an MLP with ReLU~\cite{relupaper} or Tanh activations; SIREN~\cite{sitzmann2019siren}, FINER~\cite{liu2024finer,zhu2024finer++}, WIRE~\cite{saragadam2022wire}, and Factor Fields~\cite{chen2023factor}.
We additionally test Positional Encoding (PE)~\cite{mildenhall2020nerf} and Hash Encoding (HE)~\cite{mueller2022instant} as embedding forms.
This ensures compatibility with downstream tasks derived from these models~\cite{strumpfler_img, Atzmon_2020_CVPR, chibane2020ndf, DVR, lin2020sdfsrn, yu2020pixelnerf, mildenhall2020nerf, simeonovdu2021ndf, icprainerf}.
We evaluate \methodname with Canonical-Polyadic (CP)~\cite{cphitchcock}, Tensor-Train (TT)~\cite{ttdecomp}, and Tucker (TU)~\cite{Tucker1966} decompositions modes, testing a range of ranks to control the speed vs. accuracy.

Our experiments concern several tasks: image (and video) representation, signed distance function-based geometry, and radiance fields; and a novel domain: physics simulations, where physical laws support sparse data scenarios.
We compare not only against the original INR backends but with leading task-specific models~\cite{chen2023neurbf, chen2020learning, chen2023factor, Saragadam2022DeepTensorLT, parkDeepSDFLearningContinuous2019, gropp_igr}.
While these specific methods are not \methodname compatible, they provide performance benchmarks for the field.

We use a four-layer MLP with 256 features, train for 50k steps using Adam with a learning rate of $10^{-4}$~\cite{kingma2014adam}, and report results over five runs on a NVIDIA RTX 3090.
\methodname permits extensive configurations.
The paper shows the best combination of backend and embedding. 
A total of 136 combinations are provided in the supplementary material, including additional ablations and downstream applications.

\subsection{Image Representation}
As images are second-order tensors, a simple matrix decomposition mode (CP) suffices.
In a conventional INR setup, an image is represented as $\Phi_\theta\left( x, y \right) = (r, g, b)$, where $\Phi_\theta$ denotes the network, $(x, y)$ are pixel coordinates, and $(r, g, b)$ represent the color values.
By contrast, we learn the image as $\phi_1(\mathbf{x}; \theta_1) \otimes \phi_2(\mathbf{y}; \theta_2) = (\mathbf{r}, \mathbf{g}, \mathbf{b})$, where the $\times$ operator denotes matrix multiplication.
Hence, we decompose the image into two axis-specific networks, whose outputs are combined via a matrix product.

\begin{table}[t]
  \centering
  \resizebox{\columnwidth}{!}{%
    \begin{tabular}{rl>{\centering\arraybackslash}p{0.5cm}rrrrr}
      \toprule
      & Backend                                                          & Rank & PSNR ($\uparrow$)                    & $\Delta$PSNR & SSIM ($\uparrow$)                         & T (mm:ss)  & $\Delta$T   \\
      \midrule
      \multirow{15}{*}{\rotatebox[origin=c]{90}{INR Baselines}}
      & Tanh\textsuperscript{\textdagger}                                & -    & 25.04 $\pm$ {\scriptsize 1.20}     & -            & 0.82 $\pm$ {\scriptsize 0.08} & 5:51       & -           \\
      & Tanh+PE\textsuperscript{\textdagger}                             & -    & 31.99 $\pm$ {\scriptsize 0.06}     & -            & 0.90 $\pm$ {\scriptsize 0.01} & 6:04       & -           \\
      & Tanh+HE\textsuperscript{\textdagger}                             & -    & 35.90 $\pm$ {\scriptsize 0.47}     & -            & 0.95 $\pm$ {\scriptsize 0.01} & 7:47       & -           \\
      \cmidrule{2-8}
      & ReLU\textsuperscript{\textdagger}                                & -    & 24.22 $\pm$ {\scriptsize 0.36}     & -            & 0.84 $\pm$ {\scriptsize 0.01} & 5:50       & -           \\
      & ReLU+PE\textsuperscript{\textdagger}~\cite{mildenhall2020nerf}   & -    & 34.35 $\pm$ {\scriptsize 0.31}     & -            & 0.95 $\pm$ {\scriptsize 0.01} & 5:58       & -           \\
      & ReLU+Hash (iNGP)\textsuperscript{\textdagger}~\cite{mueller2022instant} & -    & \bm 36.22 $\pm$ {\scriptsize 0.10} & -            & 0.96 $\pm$ {\scriptsize 0.01} & 7:40       & -           \\
      \cmidrule{2-8}
      & SIREN\textsuperscript{\textdagger}~\cite{sitzmann2019siren}      & -    & 28.89 $\pm$ {\scriptsize 0.02}     & -            & 0.97 $\pm$ {\scriptsize 0.00} & 3:22       & -           \\
      & WIRE~\cite{saragadam2022wire}                                    & -    & 32.06 $\pm$ {\scriptsize 0.15}     & -            & 0.87 $\pm$ {\scriptsize 0.00} & 31:28      & -           \\
      & FINER~\cite{liu2024finer,zhu2024finer++}                         & -    & 22.73 $\pm$ {\scriptsize 0.30}     & -            & 0.75 $\pm$ {\scriptsize 0.01} & 22:45      & -           \\
      & LIIF~\cite{chen2020learning}                                     & -    & 33.10 $\pm$ {\scriptsize 0.18}     & -            & 0.98 $\pm$ {\scriptsize 0.01} & *301:25      & -           \\
      & Coord-X~\cite{Liang2022CoordXAI}                    & -    & 29.70 $\pm$ {\scriptsize 1.13}     & -            & 0.80 $\pm$ {\scriptsize 0.02} & 13:05           & -           \\
      & DeepTensor~\cite{Saragadam2022DeepTensorLT}                      & -    & 30.23 $\pm$ {\scriptsize 0.20}     & -            & 0.82 $\pm$ {\scriptsize 0.01} & 2:19       & -           \\
      & NeuRBF~\cite{chen2023neurbf}                                     & -    & \sm 41.61 $\pm$ {\scriptsize 0.01} & -            & 0.95 $\pm$ {\scriptsize 0.01} & *8:22      & -           \\
      & Factor Fields~\cite{chen2023factor}                                & -    & \gm 41.82 $\pm$ {\scriptsize 0.01} & -            & 0.95 $\pm$ {\scriptsize 0.00} & *1:35      & -           \\
      \midrule
      \midrule
      \multirow{11}{*}{\rotatebox[origin=c]{90}{\methodname}}
      & WIRE                                                             & 316  & 34.39 $\pm$ {\scriptsize 0.06}     & \be{+2.33}   & 0.89 $\pm$ {\scriptsize 0.00} & 2:07       & \be{-29:21} \\
      & WIRE+PE                                                          & 316  & 35.72 $\pm$ {\scriptsize 0.04}     & \be{+3.66}   & 0.90 $\pm$ {\scriptsize 0.00} & 2:07       & \be{-29:21} \\
      & WIRE+HE                                                          & 316  & 35.11 $\pm$ {\scriptsize 0.06}     & \be{+3.05}   & 0.88 $\pm$ {\scriptsize 0.00} & 2:17       & \be{-29:06} \\
      \cmidrule{2-8}
      & SIREN                                                            & 256  & \bm 36.18 $\pm$ {\scriptsize 0.06} & \be{+7.29}   & 0.90 $\pm$ {\scriptsize 0.00} & 1:49       & \be{-01:33} \\
      & SIREN+PE                                                         & 256  & \sm 36.28 $\pm$ {\scriptsize 0.08} & \be{+7.39}   & 0.90 $\pm$ {\scriptsize 0.00} & 1:52       & \be{-01:30} \\
      & SIREN+HE                                                         & 316  & 36.02 $\pm$ {\scriptsize 0.09}     & \be{+7.13}   & 0.90 $\pm$ {\scriptsize 0.00} & 2:16       & \be{-01:06} \\
      \cmidrule{2-8}
      & FINER                                                            & 16   & 21.24 $\pm$ {\scriptsize 0.29}     & \wo{-1.49}   & 0.74 $\pm$ {\scriptsize 0.00} & 0:59       & \be{-21:46} \\
      & FINER+PE                                                         & 128  & 30.95 $\pm$ {\scriptsize 0.11}     & \be{+8.22}   & 0.83 $\pm$ {\scriptsize 0.00} & 1:33       & \be{-21:12} \\
      & FINER+HE                                                         & 64   & 25.85 $\pm$ {\scriptsize 0.32}     & \be{+3.12}   & 0.73 $\pm$ {\scriptsize 0.01} & 1:14       & \be{-21:31} \\
      \cmidrule{2-8}
      & Factor Fields                                                    & 128  & \gm 46.48 $\pm$ {\scriptsize 0.25} & \be{+4.66}   & 0.98 $\pm$ {\scriptsize 0.01} & *2:40      & \wo{+01:05} \\
      \bottomrule
    \end{tabular}%
  }
  \caption{
    \textbf{Quantitative Image Results.}
     F-INR boosts PSNR and training speed for many backbones.
     \be{Green}/\wo{red} marks improvement/degradation to the corresponding baselines.
     \textsuperscript{\textdagger}Denotes \emph{tiny-cuda-nn} implementation~\cite{tiny-cuda-nn}.
     *LIIF/NeuRBF/Factor Fields results are at convergence (10k steps).
    \vspace{-0.5cm}
  }
  \label{tab:img}
\end{table}

Following common practice~\cite{saragadam2022wire,liu2024finer,sitzmann2019siren,MIRE_2025_CVPR}, we use the \emph{parrot} image from the DIV2K dataset~\cite{div2k} as a representative benchmark.
We additionally compare against LIIF~\cite{yokota}, DeepTensor~\cite{Saragadam2022DeepTensorLT}, NeuRBF~\cite{chen2023factor}, CoordX~\cite{Liang2022CoordXAI} and Factor Fields~\cite{chen2023factor}.
All models are trained with a batch size of $2^{18}$.
Qualitative and quantitative results are presented in \Cref{fig:img-rep} and \Cref{tab:img}, respectively.
F-INR improves many tested backbones, achieving higher reconstruction fidelity (PSNR) while simultaneously accelerating training time by up to $20\times$.
Factor Fields~\cite{chen2023factor} is slower in its F-INR setup.

With similar image quality, we display speedups over native and optimized \emph{tiny-cuda-nn}~\cite{tiny-cuda-nn} implementations.
This confirms that our functional decomposition benefits are orthogonal to and compound with low-level hardware acceleration.
An optimized F-INR CUDA implementation could yield additional speedups, but is out of the scope of this study.
We further validate this versatility on single-image super-resolution, denoising, and video encoding tasks, with details provided in the supplementary material.

\subsection{Signed Distance Functions for Geometries}
\label{ssec:sdf}
A Signed Distance Function (SDF) is a continuous function assigning a signed distance to every point in 3D space relative to a surface.
While many methods learn SDFs from point clouds~\cite{sitzmann2019siren, saragadam2022wire, parkDeepSDFLearningContinuous2019}, we focus on reconstructing continuous geometries from dense voxel grids.
This structured representation is well-suited for applications requiring spatial coherence and is a challenging benchmark for continuous models~\cite{yao20213d, liu2021voxel, wu2022voxurf, Shin_2024_ACCV}.

To ensure the learned function is a valid metric space, we adopt a physics-informed learning approach by enforcing the Eikonal equation.
Our loss function, building on prior work~\cite{sitzmann2019siren, parkDeepSDFLearningContinuous2019, Park_2019_CVPR, gropp_igr}, is defined as:
\begin{align}
    \mathcal{L}_{\text{SDF}} &= \int_{\Omega }^{}\left\| \bigtriangledown \Psi \left ( x,y,z,\Theta \right ) - 1 \right\| dxdydz\\
    &+\int_{\Omega}^{} \left\| \Psi\left ( x,y,z,\Theta \right )- \hat{\Psi}(x,y,z) \right\| dxdydz  \nonumber \\ 
    &+ \int_{\Omega<\Omega_0}^{} \left\| \Psi\left (x,y,z,\Theta \right )- \hat{\Psi}(x,y,z) \right\| dxdydz,  \nonumber
\end{align}
where $\Omega$ is the spatial domain.
The loss comprises three terms: an Eikonal regularization that forces the gradient norm of the SDF to be unity, a data fidelity term matching the ground truth $\hat{\Psi}$ over the full domain, and a surface-focused term that prioritizes accuracy near the object's surface  $\Omega < \Omega_0$.
Following~\cite{Sommer_2022_CVPR}, we truncate ground-truth SDF values beyond a threshold of $0.1$.

This PDE-constrained learning paradigm places a strong requirement on the model architecture: it must be fully differentiable with respect to its input coordinates to compute the gradient $\nabla\Psi$.
Methods like TensoRF~\cite{chen2022tensorf}, Factor Fields~\cite{chen2023factor}, and NeuRBF~\cite{chen2023neurbf}, which rely on discrete lookups or factorizations, are incompatible with this objective.
Likewise, we exclude hash encoding combinations (iNGP) from our comparison~\cite{mueller2022instant, HUANG2024112760}.
The piecewise, non-smooth nature of hash grids is incompatible with this regularization requirement of gradients at the inputs.

We evaluate our method on geometries from the Stanford 3D Scan Repository~\cite{stan1, stan2, stan3, stan4}, showing the \emph{Armadillo} in the main paper and more in the supplementary material. 
The experimental protocol, including backbones and hyperparameters, is consistent with prior sections.

\begin{figure}[t]
    \centering
    \includegraphics[width=\linewidth]{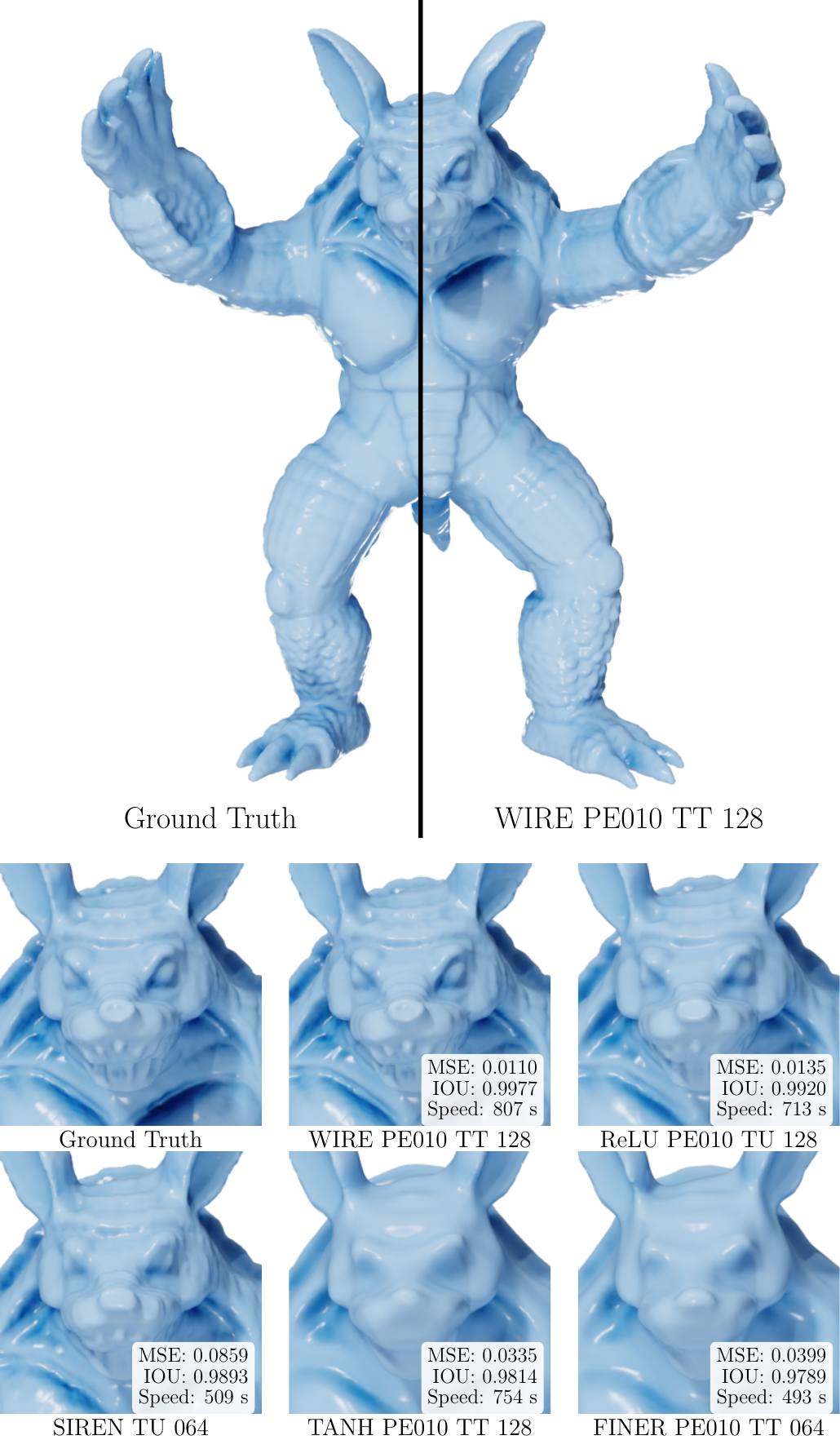}
    \caption{
        \textbf{Qualitative SDF Results.}
        For \emph{Armadillo}, F-INR models with strong inductive biases (WIRE~\cite{saragadam2022wire}, SIREN~\cite{sitzmann2019siren},  ReLU+PE~\cite{mildenhall2020nerf}) capture fine details with high fidelity.
        Others, like Tanh and FINER~\cite{liu2024finer}, preserve the macro-structure but fail to reconstruct high-frequency details, resulting in oversmoothed surfaces and lower IOU scores.
        More examples are provided in the supplementary material.
        \vspace{-0.3cm}
    }
    \label{fig:sdf_quali}
\end{figure}

As shown in \Cref{fig:sdf_quali} and \Cref{tab:arma_all}, F-INR's performance is strongly dependent on the choice of INR backbone.
Architectures with powerful inductive biases, such as SIREN~\cite{sitzmann2019siren}, WIRE~\cite{saragadam2022wire}, or ReLU+PE~\cite{mildenhall2020nerf}, excel at capturing high-frequency details like armor plating and facial wrinkles, leading to high IOU scores.
In contrast, backends like Tanh and FINER~\cite{liu2024finer,zhu2024finer++} learn the coarse geometry but have strong oversmoothing, losing fine details.
These results highlight that performance is a complex interplay between the backend, mode, and rank.
While a comprehensive search is required for optimal results, our experiments suggest emerging design principles, such as matching high-rank decompositions with powerful backbones, that can guide the efficient configuration of the framework. 

While most of the 136 tested combinations yield plausible geometries, we observe reconstruction failures in some cases (see supply. mat. Tanh+PE CP 32).
We hypothesize this occurs when the tensor rank is too low to model the complex interactions between spatial axes for a given backbone.
An extensive analysis, including full quantitative results and an ablation study on decomposition-rank stability, is provided in the supplementary material.

\begin{table}[t]
  \centering
  \resizebox{1.0\columnwidth}{!}{%
    \begin{tabular}{llcrrrr}
      \toprule
      & Backend                                          & Mode & Rank & IoU ($\uparrow$)                           & MSE ($\downarrow$)                          & T (mm:ss) \\
      \midrule
      \multirow{9}{*}{\rotatebox[origin=c]{90}{INR Baselines}}
      & ReLU                                            & -    & -    & 0.941 $\pm$ {\scriptsize 0.002} & 0.295 $\pm$ {\scriptsize 0.002} & 250:57    \\
      &ReLU+PE~\cite{mildenhall2020nerf}                & -    & -    & 0.950 $\pm$ {\scriptsize 0.001} & 0.233 $\pm$ {\scriptsize 0.002} & 257:08    \\
      &WIRE~\cite{saragadam2022wire}                    & -    & -    & 0.988 $\pm$ {\scriptsize 0.003} & 0.154 $\pm$ {\scriptsize 0.004} &  252:14   \\
      &SIREN~\cite{sitzmann2019siren}                   & -    & -    & 0.989 $\pm$ {\scriptsize 0.004} & 0.127 $\pm$ {\scriptsize 0.006} &  252:45    \\
      &FINER~\cite{liu2024finer,zhu2024finer++}         & -    & -    & 0.989 $\pm$ {\scriptsize 0.001}& 0.125 $\pm$ {\scriptsize 0.001} & 217:10 \\
      &DeepSDF~\cite{parkDeepSDFLearningContinuous2019} & -    & -    & 0.989 $\pm$ {\scriptsize 0.002} & 0.126 $\pm$ {\scriptsize 0.005} &  315:32    \\
      &IGR~\cite{gropp_igr}                             & -    & -    &\bm 0.990 $\pm$ {\scriptsize 0.004} & 0.127 $\pm$ {\scriptsize 0.004} &  292:13    \\
      
      & NeuRBF*\textsuperscript{,\textdagger}\cite{chen2023neurbf}  & -  & - & \gm 0.997 $\pm$ {\scriptsize 0.003} & 0.153 $\pm$ {\scriptsize 0.001} &  5:22    \\
      & Factor Fields\textsuperscript{\textdagger}\cite{chen2023factor}  & -  & - & 0.972 $\pm$ {\scriptsize 0.017} & $10^{-6}$ $\pm$ {\scriptsize 0.000} &  5:24    \\
      
      \midrule
      \midrule
      \multirow{5}{*}{\rotatebox[origin=c]{90}{F-INR}}
      &Tanh                                             & TT   & 128  & 0.941 $\pm$ {\scriptsize 0.002} & 0.034 $\pm$ {\scriptsize 0.004} & 12:34      \\
      &ReLU+PE                                          & TU   & 128  & \sm 0.992 $\pm$ {\scriptsize 0.001} & 0.013 $\pm$ {\scriptsize 0.002} & 11:54      \\
      &WIRE+PE                                          & TT   & 128  & \gm 0.997 $\pm$ {\scriptsize 0.002} & 0.011 $\pm$ {\scriptsize 0.003} & 13:27      \\
      &SIREN                                            & TU   & 64   & 0.989 $\pm$ {\scriptsize 0.003} & 0.085 $\pm$ {\scriptsize 0.006} & 8:29      \\
      &FINER+PE                                         & TT   & 64   & 0.978 $\pm$ {\scriptsize 0.002} & 0.039 $\pm$ {\scriptsize 0.002} & 8:13      \\
      \bottomrule
    \end{tabular}%
  }
  \caption{
    \textbf{Quantitative SDF Results.}
    F-INR outperforms both the original INR backbones and specialized methods like DeepSDF and IGR on \emph{Armadillo}.
    It simultaneously improves geometric fidelity (higher IoU, lower MSE) while accelerating training by over $20\times$.
    \textdagger~ denotes models trained without Eikonal PDE regularization.
    Results for all combinations are in the supplementary material. 
    *Evaluated on an \emph{RTX 6000 Ada} due to memory constraints.
  }
  \label{tab:arma_all}
\end{table}
\subsection{Neural Radiance Fields}
Neural Radiance Fields (NeRFs) represent a 3D scene as a continuous function that maps a 5D input: a 3D coordinate $(x, y, z)$ and a 2D viewing direction $(\theta, \phi)$, to a 4D output of color and volume density.
A model is optimized by querying points sampled along rays cast from known camera poses to reconstruct a set of training images.
The goal is to render photorealistic novel views by interpolating the learned function across unseen camera viewpoints.
Prior work improves NeRF through architectural changes, embeddings, or optimized sampling strategies~\cite{mildenhall2020nerf, mueller2022instant, icprainerf, sitzmann2019siren, Liang2022CoordXAI}.
In contrast, we reformulate the underlying functional representation.

This raises a crucial question: How can an axis-aligned functional decomposition operate effectively on the sparse, non-grid coordinate queries inherent to NeRF?
Our functional decomposition operates on the input coordinate domains, not on a pre-defined grid of samples.
For a NeRF function $f(x, y, z, \theta, \phi)$, F-INR factorizes it into a combination of five low-dimensional sub-networks, each modeling a single coordinate axis (e.g., $f_x(x), f_y(y), \dots$).
The closest work would be TensorRF~\cite{chen2022tensorf}, which learns low-rank representations, but does not use per-axis MLPs and depends on tri-linear interpolation for continuity, and it is not modular.
This factorization remains valid even for unstructured input coordinates but increases per-iteration training time.
We hypothesize that the sub-network representation leads to reconstruction fidelity improvements.

\begin{figure*}[t]
    \centering
    \includegraphics[width=\linewidth]{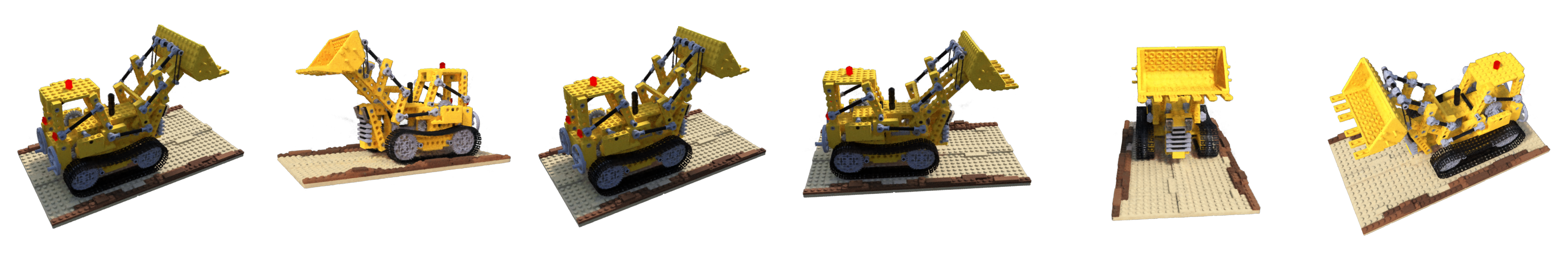}
    \includegraphics[width=\linewidth]{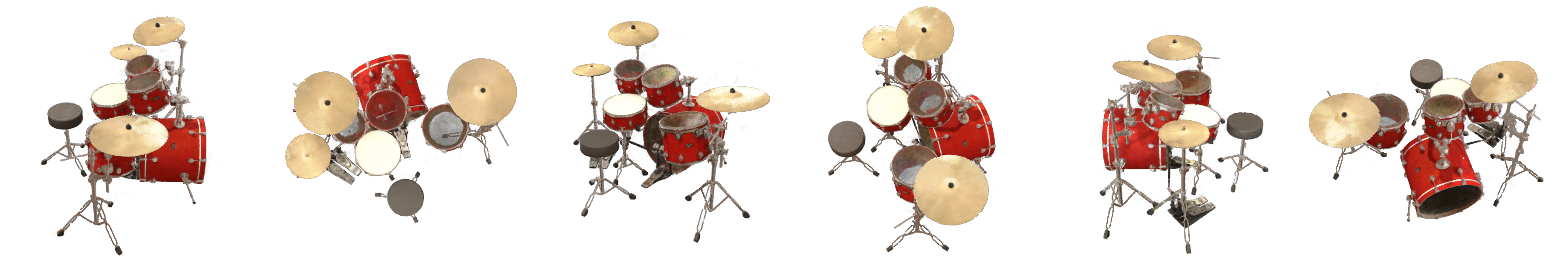}
    \caption{
        \textbf{Qualitative NeRF Results.}
        Novel view renderings from our best-performing F-INR model (ReLU+HE with a TT rank of 16) on the \emph{Lego} (top) and \emph{Drums} (bottom) scenes.
        The results demonstrate the model's ability to reconstruct fine geometric details and complex view-dependent effects with high fidelity.
        \vspace{-0.2cm}
    }
    \label{fig:nerf}
\end{figure*}

We follow the experimental protocol from~\cite{mildenhall2020nerf} and evaluate on the synthetic dataset.
We present quantitative and qualitative results for the \emph{Lego} and \emph{Drums} scenes in the main paper and all remaining scenes in the supplementary material.
We report PSNR on the validation set in \Cref{tab:nerf} and provide six synthesized novel views in \cref{fig:nerf}.

\begin{table}[t]
    \centering
    \resizebox{\columnwidth}{!}{%
    \begin{tabular}{llrrrrr}
    \toprule
     & Backend & Mode & Rank & Lego & Drums & T  \\
    \midrule
    \multirow{8}{*}{\rotatebox[origin=c]{90}{INR Baselines}}
       &NeRF~\cite{mildenhall2020nerf}       & - & - &    30.83 &     22.01 & 35 hours \\
       &DVGO~\cite{sun2022direct}            & - & - &    31.25 &     24.90 & 24.4 min \\
       &Plenoxels~\cite{yu2021plenoxels}     & - & - &    31.71 &     24.48 & 22.5 min \\
       &TensoRF~\cite{chen2022tensorf}       & - & - &    33.14 &     25.26 & 25.8 min \\  
       &ReLU+Hash (iNGP)~\cite{mueller2022instant} & - & - &    34.59 &     25.28 &  4.8 min \\
       &NeuRBF*~\cite{chen2023neurbf}        & - & - &\gm 37.29 & \gm 26.57 & 1.1 hours \\ 
       &Factor Fields~\cite{chen2023factor}  & - & - &    33.14 & \gm 26.57 & 28.4 min \\ 
       &K-Planes~\cite{kplanes_2023}  & - & - &    34.23 &  \bm 25.30 & 76.8 min \\
    \midrule
    \midrule
    \multirow{6}{*}{\rotatebox[origin=c]{90}{F-INR}}
       &ReLU+HE                              & CP &  4 &    31.87 &     24.55 &  5.8 min \\
       &ReLU+HE                              & CP &  8 &    33.88 &     25.12 &  8.1 min \\
       &ReLU+HE                              & CP & 16 &\bm 34.81 & \bm 25.30 & 18.3 min \\ 
       &ReLU+HE                              & TT &  8 &    34.24 &     25.07 & 16.2 min \\
       &ReLU+HE                              & TT & 16 &\sm 35.67 & \sm 26.48 & 22.4 min \\
       &ReLU+HE                              & TU &  8 &    31.22 &     23.02 & 28.6 min \\
   \bottomrule
    \end{tabular}%
    }
    \caption{\textbf{Quantitative NeRF Results.}
    We show the PSNR values for the \emph{Lego} and \emph{Drums} scenes of the synthetic NeRF dataset~\cite{mildenhall2020nerf}.
    F-INR performs best with hash encoding (HE), and Tensor Train (TT) outperforms CP and TU.
    We focus on variants using a ReLU+HE backbone to best illustrate these findings.
    *Evaluated on an \emph{RTX 6000 Ada} due to memory constraints.
    }
    \label{tab:nerf}
\end{table}

This experimental setup demonstrates how tensorized univariate MLPs enable faster and more accurate solutions. We adopt the state-of-the-art Hash encoding-based NeRF~\cite{mueller2022instant}, which uses 3D hash grid encodings. 
In our approach, we replace this with three independent 1D hash encodings while keeping all other components unchanged. 
This decomposition reduces complexity and allows \methodname-enhanced NeRF models to operate with significantly higher feature dimensions, corresponding to the tensor decomposition rank. 
A higher rank enhances the model's expressiveness in capturing finer details and resolving ambiguities more effectively, thereby improving reconstruction quality and PSNR. 
Importantly, our goal is not to outperform NeRF baselines, but to demonstrate that NeRFs can be represented within the \methodname framework. 
Although our implementation is in pure PyTorch~\cite{paszke2019pytorchimperativestylehighperformance} and lacks CUDA-specific optimizations like InstantNGP, we observe orthogonal performance gains, suggesting that a CUDA-optimized \methodname could achieve even higher training efficiency.

\subsection{Physics-Informed Learning as INR Problem}
\label{ssec:pinns}
We propose that Implicit Neural Representations (INRs) can model systems governed by physical laws, a paradigm where a physical system is a continuous function constrained by a PDE.
While \Cref{ssec:sdf} utilizes a PDE for regularization, here we investigate the more challenging setting of learning from sparse data points where a data-driven solution would be intractable.
This perspective, however, reveals a fundamental bottleneck: the monolithic architectures used in methods like Physics-Informed Neural Networks (PINNs)~\cite{Raissi2019, Wangntk, vemuri2024} struggle to efficiently represent complex, high-dimensional physical phenomena~\cite{krishnapriyan2021characterizing, WANG2022110768}.

F-INR is suited to address this representational issue.
Factorizing the high-dimensional spatio-temporal function into a product of simple, low-dimensional sub-networks decouples the system's complexity.
This idea facilitates a powerful application: learning a continuous, high-resolution physical field from only sparse, low-resolution data, using the PDE loss to enforce physical consistency everywhere else.
We hypothesize that F-INR could achieve high-fidelity, zero-shot super-resolution on tasks where monolithic models would be expensive or fail to converge.

To validate this hypothesis, we select a canonical and challenging benchmark: modeling the decaying turbulence of a 2D fluid.
This system is governed by the vorticity form of the incompressible Navier-Stokes equations~\cite{wang2022respecting}:

\begin{equation}
    \begin{array}{r@{~}lr}
        \partial_t \omega+u \cdot \nabla \omega 
        &=\nu \Delta \omega, 
        & x \in \Omega, t \in \Gamma,
        \\
        \nabla u
        &=0, 
        & x \in \Omega, t \in \Gamma, 
        \\
        \omega(x, 0)
        &=\omega_0(x), 
        & x \in \Omega,
    \end{array}
    \label{eq:ns}
\end{equation}

where $u$ denotes the velocity field, $\omega = \nabla \times u$ is the vorticity, $\omega_0$ represents the initial vorticity, and $\nu = 0.01$ is the viscosity coefficient.
The spatial domain is $\Omega \in [0, 2\pi]^2$, and the temporal domain is $\Gamma \in [0, 1]$.

\begin{table}[t]
	\centering
    \resizebox{\columnwidth}{!}{%
	\begin{tabular}{llrrrr}
		\toprule
		& Backend                              & Mode        & Rank         & MSE ($\downarrow$)                             & T (hh:mm)  \\
		\midrule
        \multirow{7}{*}{\rotatebox[origin=c]{90}{Baselines}}
		& ReLU+PE~\cite{mildenhall2020nerf}    & -           & -            & 0.097 $\pm$ {\scriptsize 0.009}          & 20:30          \\
		& WIRE~\cite{saragadam2022wire}        & -           & -            & 0.073 $\pm$ {\scriptsize 0.004}          & 20:25          \\
		& SIREN~\cite{sitzmann2019siren}       & -           & -            & 0.184 $\pm$ {\scriptsize 0.010}          & 20:24          \\
		\cmidrule{2-6}
        & ModifiedPINN~\cite{WANG2022110768}   & -           & -            & 0.074 $\pm$ {\scriptsize 0.008}          & 28:40          \\
		& CausalPINN~\cite{wang2022respecting} & -           & -            & 0.070 $\pm$ {\scriptsize 0.011}          & 33:12          \\
		& MFF Net~\cite{jiangchiu2020}         & -           & -            & 0.048 $\pm$ {\scriptsize 0.003}          & 35:18          \\
		& PhySR~\cite{REN2023112438}           & -           & -            & 0.038 $\pm$ {\scriptsize 0.020}          & 27:05          \\
		\midrule
            \midrule
        \multirow{11}{*}{\rotatebox[origin=c]{90}{F-INR}}
		&WIRE                                 & TT          & 64           & 0.034 $\pm$ {\scriptsize 0.004}          & 00:59          \\
		&WIRE                                 & TT          & 128          & 0.036 $\pm$ {\scriptsize 0.002}          & 01:08          \\
		&WIRE                                 & TT          & 256          & \bm 0.033 $\pm$ {\scriptsize 0.002}          & 01:40          \\
		&WIRE                                 & TU      & 128          & 0.037 $\pm$ {\scriptsize 0.003}          & 01:28          \\
		&WIRE & TU      & 256          & 0.035 $\pm$ {\scriptsize 0.004}          & 02:01          \\
	\cmidrule{2-6}	
        &ReLU+PE                              & TT          & 64           & \bm 0.033 $\pm$ {\scriptsize 0.001}          & 00:51          \\
		&ReLU+PE                     & TT & 128 & \gm 0.030 $\pm$ {\scriptsize 0.002} & 01:12 \\
		&ReLU+PE                     & TT & 256 & \gm 0.030 $\pm$ {\scriptsize 0.002} & 01:59 \\
		&ReLU+PE                              & TU      & 64           & 0.039 $\pm$ {\scriptsize 0.005}          & 00:45          \\
		&ReLU+PE                              & TU      & 128          & \sm 0.032 $\pm$ {\scriptsize 0.004}          & 01:34          \\
		&ReLU+PE                              & TU      & 256          & \sm 0.032 $\pm$ {\scriptsize 0.005}          & 02:03          \\
		\bottomrule
	\end{tabular}%
    }
	\caption{
        \textbf{Quantitative PINN INR Results.}
        F-INR outperforms standard INRs and specialized physics-informed baselines.
        It achieves a state-of-the-art error (0.030 MSE), while accelerating training by over $20\times$.
        The table presents top-performing F-INR configurations; full results are in the supplementary material.
        \vspace{-0.6cm}
	}
	\label{tab:ns}
\end{table}

The ground-truth simulation has a spatio-temporal resolution of $101\times128\times128$~\cite{wang2022respecting}.
For training, a model only observes a coarse, downsampled version at a $10\times64\times64$ resolution, a 40-fold reduction in data points.

The model optimizes a composite loss: $\mathcal{L} = \mathcal{L}_{\text{data}} + \mathcal{L}_{\text{pde}}$, 
where $\mathcal{L}_{\text{data}}$ is the $L_2$ norm between the prediction and the coarse grid data, and $\mathcal{L}_{\text{pde}}$ is the MSE of the PDE residual from \Cref{eq:ns}, sampled at domain collocation points.
Our framework learns a single, continuous function over the entire spatio-temporal domain.
This holistic representation avoids the need for complex training strategies like time-marching~\cite{krishnapriyan2021characterizing}, used to stabilize PINNs on long sequences.

\begin{figure}[t]
    \centering
    \includegraphics[width=\linewidth]{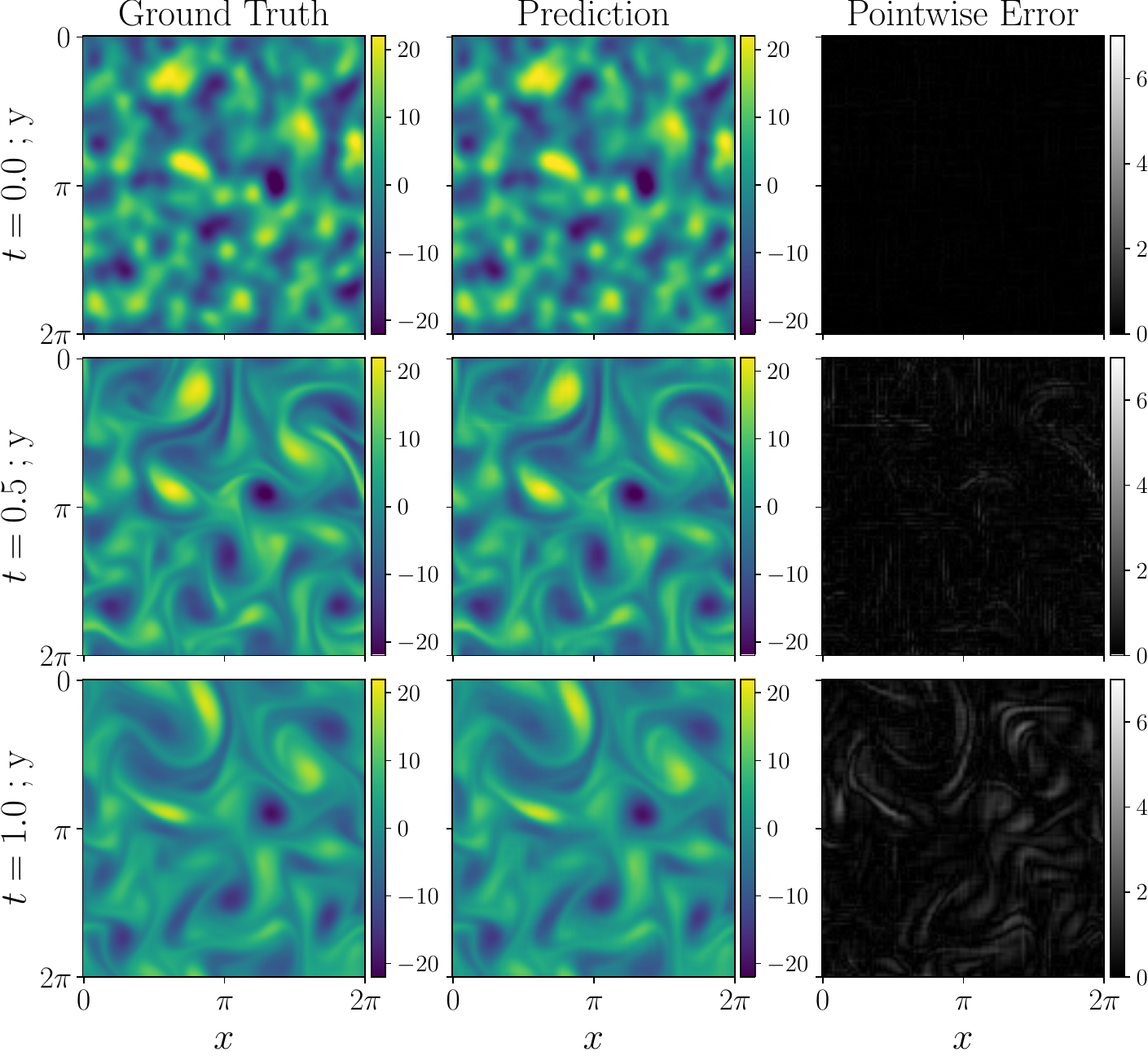}
    \caption{
        \textbf{Qualitative PINN INR Results.}
        We compare the vorticity field (left) against our F-INR model (ReLU+PE TT rank 128) prediction (middle) and visualize the pointwise absolute error (right) at three steps.
        Trained on sparse data, the model captures the complex dynamics across the entire temporal domain.
        We provide the full simulation video in the supplementary material.
        \vspace{-0.3cm}
    }
    \label{fig:ns}
\end{figure}

Architectures that employ discrete lookups or non-smooth representations (e.g., TensoRF, hash grids, Factor Fields) are fundamentally ill-suited for physics-informed objectives that depend on stable, high-order gradient computation.
As such, they are omitted from this comparison.

The quantitative results in \Cref{tab:ns} demonstrate that F-INR achieves a new state-of-the-art in physics-informed super-resolution.
We test against two baseline forms: general-purpose INRs adapted for PINNs, and specialized architectures designed for this task, such as CausalPINN~\cite{wang2022respecting}, ModifiedPINN~\cite{WANG2022110768}, MeshFreeFlowNet~\cite{jiangchiu2020}, and PhySR~\cite{REN2023112438}.
Our best F-INR configuration (ReLU+PE, TT, rank 128) surpasses all baselines, achieving an MSE of $0.030$ compared to the previous best of $0.038$ (PhySR).
More importantly, it achieves this with better accuracy while reducing training time by over 20$\times$, from 27 hours to under two.
This highlights F-INR's ability to overcome the established speed-accuracy trade-off.
Notably, our results represent the first successful application of the WIRE architecture~\cite{saragadam2022wire} to a complex physics simulation, empirically showing the versatility of our framework.

\Cref{fig:ns} depicts turbulent flow reconstruction.
The error map is not uniform; it focuses on high-gradient areas (vortex edges) and increases as the turbulence develops.
We attribute this to the challenge of capturing the high-frequency spectral content inherent to dynamic features.
An analysis of model capacity (backend and rank) is provided in the supplementary material via several sparsity ablations.

\section{Limitations}
Methodologically, we concentrated on established tensor formats.
However, exploring more advanced decompositions, such as Tensor Rings (TR)~\cite{zhao2016tensor}, could offer more substantial compression-accuracy trade-offs per backend, especially for higher-dimensional tasks, such as NeRFs.

The introduced F-INR combinatorial space invites the development of automated techniques, such as adaptive rank selection, to move beyond grid searches.
Nevertheless, we provide first selection rules from our experiment observations in the supplementary material.

On the engineering side, a dedicated and optimized CUDA implementation integrated into a library like \emph{tiny-cuda-nn}~\cite{tiny-cuda-nn} presents a clear path to substantial, further performance gains, but was out of scope for this study.

Our work presents an empirically successful solution to the curse of dimensionality~\cite{HORNIK1991251,GRUNE2021317,Poggio2017}, a well-known theoretical challenge for high-dimensional function approximation.
A formal theory that explains why functional decomposition with neural network is effective, however, remains an open research area.
Such an analysis needs to characterize the relationship between tensor structure, expressive power, and the optimization dynamics of factorized INRs.

\section{Conclusion}
We introduce F-INR, a framework that reframes high-dimensional implicit neural representation through the lens of functional tensor decomposition. 
By factorizing a monolithic network into a product of compact, axis-aligned sub-networks, F-INR provides a new direction that is orthogonal to network design. 
This approach decouples the learning problem, mitigating the curse of dimensionality.

Our experiments validate this paradigm across a wide range of INR tasks, from canonical image representation to scientific applications, such as PDE-based super-resolution.
This factorization yields training accelerations of up to $20\times$ over native backends and provides speedups compared to highly optimized CUDA kernels, demonstrating fundamental benefits.
By operating on the functional coordinate space itself, our framework seamlessly handles both grid-aligned and unstructured, query-based data without requiring additional data interpolation.

Our work opens several research directions, including adaptive rank selection, advanced tensor formats, and dedicated hardware acceleration.
However, we believe the most significant impact of this work is conceptual.
It advocates a perspective that is broadly applicable: that improvements in machine learning can be unlocked not just by building better models, but by restructuring the functions they are tasked with learning.
This functional factorization principle may be a powerful and transferable paradigm for complex learning problems in other domains.

\newpage
{
\small
\noindent\textbf{Acknowledgment}
A great number of people have contributed to this paper, whether consciously or not.
We especially thank the members of our group for their contributions throughout the revisions of this manuscript, which ultimately brought it to a publication-worthy state.
}

{
    \small
    \bibliographystyle{ieeenat_fullname}
    \bibliography{main}
}

\end{document}